\pdfoutput=1

\documentclass[11pt]{article}

\usepackage[preprint]{acl}

\usepackage{times}
\usepackage{latexsym}

\usepackage[T1]{fontenc}

\usepackage[utf8]{inputenc}

\usepackage{microtype}

\usepackage{inconsolata}

\usepackage{graphicx}

\usepackage{xprintlen}
\usepackage{enumitem}
\usepackage[acronym]{glossaries}
\usepackage{lipsum}
\usepackage{xcolor}
\usepackage[most]{tcolorbox}
\usepackage{float}

\title{OpenReviewer: A Specialized Large Language Model for Generating Critical Scientific Paper Reviews}

\author{Maximilian Idahl \\
  L3S Research Center\\
  Leibniz University Hannover \\
  \texttt{idahl@l3s.de} \\
  \And
  Zahra Ahmadi \\
  PLRI Medical Informatics Institute \\
  Hannover Medical School \\
  \texttt{ahmadi.zahra@mh-hannover.de} \\
}

\begin{document}
\maketitle
\begin{abstract}
We present OpenReviewer, an open-source system for generating high-quality peer reviews of machine learning and AI conference papers. At its core is Llama-OpenReviewer-8B\footnote{Model: \href{https://huggingface.co/maxidl/Llama-OpenReviewer-8B}{huggingface.co/maxidl/Llama-OpenReviewer-8B}}, an 8B parameter language model specifically fine-tuned on 79,000 expert reviews from top conferences. Given a PDF paper submission and review template as input, OpenReviewer extracts the full text, including technical content like equations and tables, and generates a structured review following conference-specific guidelines. Our evaluation on 400 test papers shows that OpenReviewer produces considerably more critical and realistic reviews compared to general-purpose LLMs like GPT-4 and Claude-3.5. While other LLMs tend toward overly positive assessments, OpenReviewer's recommendations closely match the distribution of human reviewer ratings. The system provides authors with rapid, constructive feedback to improve their manuscripts before submission, though it is not intended to replace human peer review. OpenReviewer is available as an online demo\footnote{Demo: \href{https://huggingface.co/spaces/maxidl/openreviewer}{huggingface.co/spaces/maxidl/openreviewer}} and open-source tool.
\end{abstract}

\section{Introduction}
The peer review process is fundamental to maintaining scientific rigor in academic research, particularly in fast-moving fields like machine learning (ML) and artificial intelligence (AI).
As submission volumes to major conferences continue to surge -- with top venues receiving over 10,000 submissions annually -- the traditional peer review system faces challenges.
The task load for reviewers consistently increases, while authors lack access to preliminary expert feedback with a quick turnaround that could help improve their work before submission.

Large language models (LLMs) have recently demonstrated remarkable capabilities in understanding and generating academic content, suggesting their potential to assist in peer review~\citep{what_can_nlp_do_for_peer_review}.
However, generating high-quality reviews presents unique challenges beyond general language understanding. A good reviewer has to comprehend technical content, including mathematics and empirical results, and evaluate the work's contributions, novelty, significance, and methodological soundness according to high, field-specific standards.

In this paper, we present OpenReviewer, an open-source system designed to generate human-like, high-quality reviews of machine learning and AI papers.
At its core is Llama-OpenReviewer-8B, a specialized language model fine-tuned on a curated dataset of 79,000 reviews from top ML conferences.
Unlike general-purpose LLMs, OpenReviewer is trained to follow standard review templates and guidelines, ensuring structured, critical feedback that aligns with conference reviewing practices.

Our main contributions are:
\begin{itemize}
\item A specialized long-context large language model for generating academic reviews, finetuned on a large dataset of expert reviews from top ML conferences.
\item An open-source demo that combines state-of-the-art transformer-based PDF processing with our specialized model to generate comprehensive reviews from paper submissions.
\item An evaluation demonstrating that OpenReviewer generates reviews that align considerably better with human expert reviews compared to state-of-the-art LLMs, including GPT-4o and Claude-3.5-Sonnet.
We find that general-purpose LLMs are not critical enough and tend to give much more positive recommendations than human reviews.
\end{itemize}

While OpenReviewer is not intended to replace human peer reviews, it provides authors with a valuable tool for obtaining rapid, structured feedback before submission. Our evaluation shows that OpenReviewer's reviews closely match human reviewer judgments, suggesting its potential to help authors identify and address weaknesses in their manuscripts in the writing process.

\section{Motivation}
The motivation behind OpenReviewer is not to replace human peer reviews.
Instead, we want to assist authors who face challenges in the pre-submission phase.
Without access to expert feedback before submission, they may overlook significant weaknesses in their manuscripts or fail to address potential reviewer concerns.
This can result in unnecessary desk rejections or negative reviews that could have been avoided with earlier feedback.

While recent advances in large language models have shown promise in various tasks related to academic research, including paper summarization, understanding, and analysis, existing models often struggle to generate reviews that match the depth, specificity, critical perspective, and structure expected in academic peer review. General purpose LLMs may miss field-specific conventions, fail to properly evaluate technical contributions, or provide feedback that does not align with established reviewing practices.
OpenReviewer addresses this gap, aiming to provide authors with valuable pre-submission feedback that closely mirrors the standards and expectations of human peer reviewers.

\section{OpenReviewer}

\subsection{Demo Interface}
\begin{figure*}[h]
    \centering
    \frame{\includegraphics[width=\textwidth]{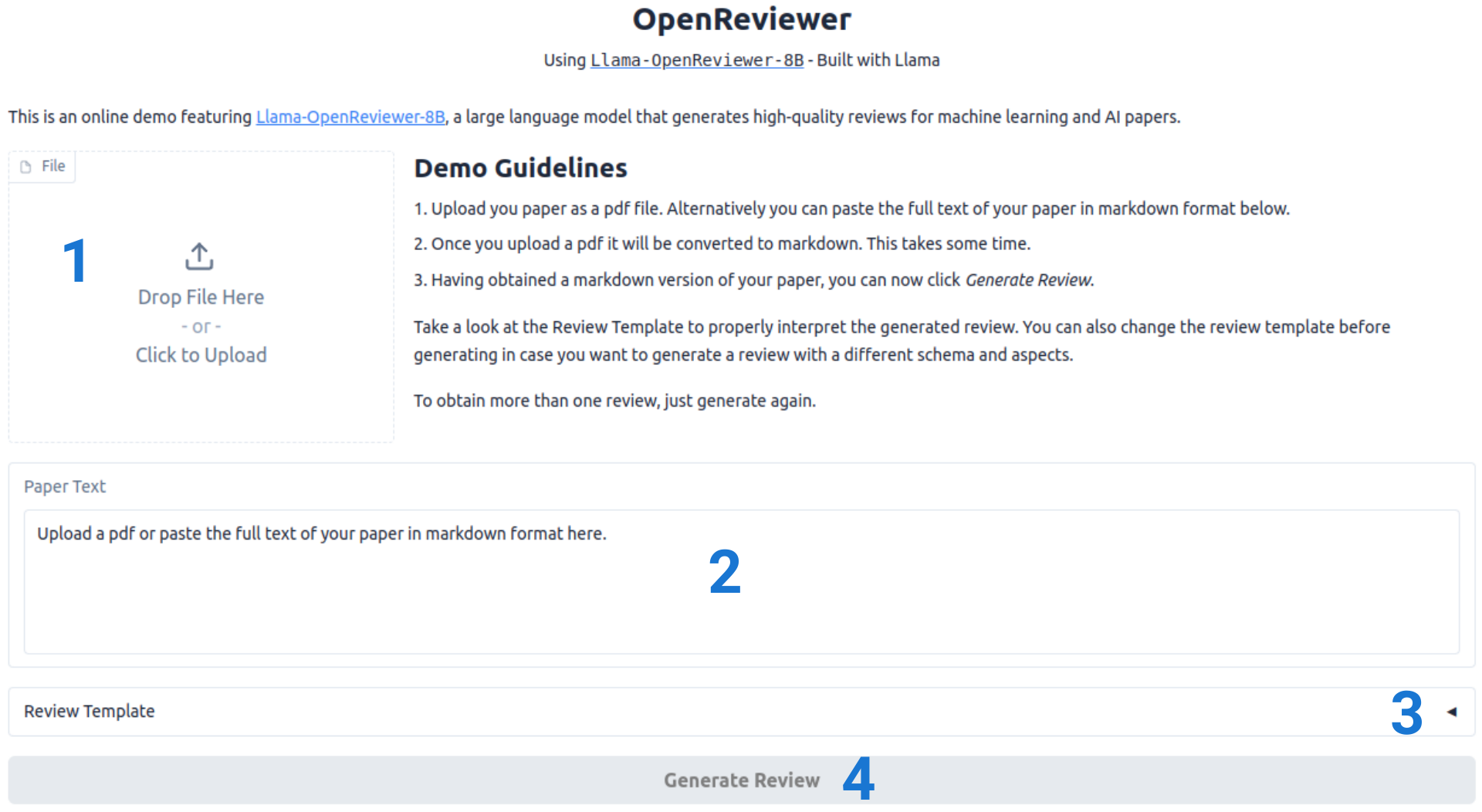}}
    \caption{Annotated screenshot of the \href{https://huggingface.co/spaces/maxidl/openreviewer}{OpenReviewer demo hosted on Huggingface Spaces}, with slightly modified layout. \textcolor[HTML]{0D47A1}{1)} Dialogue for uploading a PDF file.
    \textcolor[HTML]{0D47A1}{2)} Once the user uploads a file, this text field will be populated with the papers' full text in markdown format. The user can choose to edit the text to fix conversion errors.
    \textcolor[HTML]{0D47A1}{3)} An accordion element to show and optionally edit the review template used for generation.
    \textcolor[HTML]{0D47A1}{4)} Button to run the review generation and enabled once the paper text field is populated. When clicked, a review is generated in streaming mode and printed below on the fly.}
    \label{fig:screen}
\end{figure*}
We host a demo of OpenReviewer on HuggingFace Spaces\footnote{\href{https://huggingface.co/spaces}{huggingface.co/spaces}}.
The interface, depicted in Figure~\ref{fig:screen}, is built with the Gradio~\citep{gradio} library.
It starts with a short description and some guideline text to help users navigate the application.
The user first faces a file-upload dialogue, where they can upload a PDF file.
Uploading a file automatically triggers a PDF to markdown conversion process.
This process takes some time, as OpenReviewer uses transformer-based PDF processing models that run on a GPU.
Once the markdown conversion finishes, the markdown paper text will be displayed in a corresponding text area.
The user can edit the text to fix any conversion errors.
Alternatively, if the user already has a markdown representation of their paper, they can avoid the markdown conversion by directly pasting it into this text area.
Below the text area for the paper text, there is an accordion element that optionally shows an editable review template.
The reviews generated by OpenReviewer follow this template, and the user can choose to deviate from the default template if they desire a review with different sections, aspects, or rating scales.
Once the markdown paper text field is populated, the user can click a ``Generate Review'' button to run Llama-OpenReviewer-8B.
When clicked, a review is generated in streaming mode and printed on the fly, token-by-token, below the button.

The demo uses Huggingface Spaces ZeroGPU hardware\footnote{\href{https://huggingface.co/docs/hub/spaces-zerogpu}{huggingface.co/docs/hub/spaces-zerogpu}}, which allocates and releases GPU resources dynamically.
While this lets us host the demo free of charge up to a particular usage quota, it can negatively impact its speed and snappiness.
However, users can clone and run the application on their hardware at any time if desired.

\subsection{Llama-OpenReviewer-8B}
The core component powering OpenReviewer is Llama-OpenReviewer-8B, a large language model finetuned on a large dataset of paper-review pairs.
This section describes the data collection, prompt design, and training details.

\subsubsection{Training Data}
From OpenReview\footnote{\href{https://openreview.net/}{openreview.net}}, we collected a dataset of 36K submitted papers and 141K reviews from the International Conference on Learning Representations (ICLR) and the Conference on Neural Information Processing Systems (NeurIPS), considering editions from 2022 onwards.
We obtain each paper in PDF format by downloading the earliest revision possible.
Later revisions are typically camera-ready versions that already incorporate feedback from the reviews, rendering at least some parts of the reviews invalid.
Unfortunately, the original double-blind submissions are no longer available for some venues; we obtain the non-anonymized version.
We convert all papers from PDF to markdown using Marker\footnote{\href{https://github.com/VikParuchuri/marker}{github.com/VikParuchuri/marker}}. This open-source PDF processing pipeline combines heuristics with multiple transformer models to extract text, apply optical character recognition (OCR) when necessary, detect page layouts, determine the reading order, clean and format text blocks, combine blocks of text, and post-process complete text.
We chose Marker because its per-page accuracy improves upon Nougat~\citep{nougat}. It can also convert equations and tables accurately, which we deem essential for scientific papers.
We discard any appendix content and only retain the full text of the main and reference sections.
We filter papers and reviews by length, removing the top and bottom 1\% quantile.
Finally, we keep only high-confidence reviews.
For each venue, we select a reviewer confidence threshold roughly equal to ``Confident, but not absolutely certain''.
After filtering, approximately 79K reviews are remaining.

\subsection{Prompt Design}
OpenReviewer uses a system prompt that conditions the LLM on its reviewer role and defines a fixed set of reviewer guidelines inspired by the ICLR 2024 reviewer guide\footnote{iclr.cc/Conferences/2024/ReviewerGuide}.
The system prompt specifies that the review must be written in markdown format and follow a specific review template, which is part of the input and differs across venues.
The user prompt is minimalistic and only contains ``Review the following paper:'' followed by the full paper text.
The verbatim prompts are shown in Section~\ref{sec:prompts}.

\subsection{Training}
We full finetune Llama-3.1-8B-Instruct for three epochs with an effective batch size of 64 and a learning rate of $2\times10^{-5}$, using bfloat16 precision.
The maximum sequence length during finetuning is 128k tokens to accommodate long paper texts.
Model training used the axolotl\footnote{axolotl.ai} library using Deepspeed ZeRO-3~\citep{deepspeed} for parallelization and Flash Attention V2~\citep{flashattentionv2} and LIGER Kernel~\citep{ligerkernel} to reduce memory usage and increase throughput.
The training process took approximately 34 hours using 64 NVIDIA A100 80GB GPUs.
Refer to Section~\ref{sec:hyperparams} for all training hyperparameters.

\section{Evaluation}
Evaluating automatically generated paper reviews is challenging.
There is no single objective measure of review quality since even human experts often disagree in their assessments.
Additionally, evaluating free-form generated text presents a challenge because there are many valid ways to express the same content.
Unlike tasks with clear right or wrong answers, free-form text can vary greatly in style, structure, word choice, and level of detail while still being equally valid or effective. For example, two reviewers could make the same core criticism about a paper's methodology but express it differently. One reviewer might be more direct and concise, while another could be more elaborate with detailed examples.
This makes it difficult to develop automated metrics that can reliably assess the quality of generated review text.

Our evaluation approach for OpenReviewer is based on comparing generated reviews to human reviews.
This approach assumes that similarity to human reviews equals quality, which may not always be accurate as the quality control for human-written reviews is limited.

To measure how well OpenReviewer reviews align with reviews from human reviewers, we conduct experiments using a test set of 400 held-out papers and their reviews from NeurIPS 2024 and ICLR 2025, the most recent venues in our dataset.
We compare OpenReviewer to Llama-3.1-8B-Instruct and Llama-3.1-70B-Instruct\cite{llama31}, Claude-3.5-Sonnet (Oct. 22) from Anthropic, and GPT-4o (2024-11-20) from OpenAI.
We generate one review for each paper in the test set using greedy decoding (temperature of 0).
All LLMs are instructed with the same system and user prompts used by OpenReviewer.
We use vLLM~\citep{vllm} to serve the OpenReviewer and Llama models, and access Claude-3.5-Sonnet and GPT-4o via OpenRouter\footnote{openrouter.ai/}.

\subsection{Matching Recommendations}
While the sections in a review vary based on the corresponding venue review template, the recommendation is a numerical rating that consistently exists for all reviews.
Additionally, the recommendation can be expected to reflect the overall sentiment and arguments expressed throughout the review.
To measure how well the recommendation of a generated review matches the recommendations of the human reviewers, we check whether it exactly matches one of the human reviewers' recommendations.
Additionally, we measure the average absolute distance between the generated review's recommendation and the human reviewers' average recommendations.
For this, we normalize the recommendation scores to a scale from 1 (strong reject) to 10 (strong accept).
As shown in Table~\ref{tab:recommendation_measures}, the recommendations by OpenReviewer match the human
reviewers much better than the other LLMs across both metrics.
OpenReviewer matches at least one human reviewer for $55.5\%$ of its generated reviews and has an average recommendation error of $0.96$. In contrast, GPT4o matches at least one reviewer only in $23.8\%$ of reviews with a higher average recommendation error of $2.34$.

\begin{table}[h]
  \centering
  \resizebox{\columnwidth}{!}{
  \begin{tabular}{lcc}
    \hline
    \textbf{Model} & \textbf{EM (\%)} & \textbf{{Avg. Error (\%)}} \\
    \hline
    Llama-3.1-8B-Instruct     & $14.0$ & $2.95\pm1.19$    \\
    Llama-3.1-70B-Instruct     & $11.5$ & $3.03\pm1.34$     \\
    Claude-3.5-Sonnet     &      $15.5$    & $2.77\pm1.27$  \\
    GPT-4o (2024-11-20)      & $23.8$       & $2.34\pm1.17$   \\
    OpenReviewer     & \underline{$55.5$}     & \underline{$0.96\pm0.85$}    \\
    \hline
  \end{tabular}}
  \caption{Exact Match (EM) and average error for the recommendation in 400 test reviews generated with different LLMs and normalized to a scale from 1 (strong reject) to 10 (strong accept).
  EM measures how often the LLM's recommendation matches at least one of the human reviews. The average error is the average absolute difference between the LLM's recommendations and the human reviewers' average recommendation. The recommendations by OpenReviewer match the human reviewers much better than other state-of-the-art LLMs.}
  \label{tab:recommendation_measures}
\end{table}

When examining why such a large gap exists between OpenReviewer and all other LLMs, we find that the other LLMs usually give positive recommendations (Table~\ref{tab:recommendation_avg}).
While OpenReviewer matches the human reviewers with an average recommendation of 5.4 out of 10, the baseline LLMs produce average recommendations of 6.9 and higher, topped by Llama-3.1-8B-Instruct with an average recommendation of 8.1, which would lead to an ``accept'' for most of the papers in the test dataset.
Although a positive recommendation can make authors happy, it is less desirable if the authors seek critical feedback for their paper pre-submission.
Recommendation scores that accurately align with human reviewer judgments provide authors with realistic expectations about their manuscript's reception and help prevent the disappointment that might follow from overly optimistic preliminary reviews.

\begin{table}[h]
  \centering
  \resizebox{\columnwidth}{!}{
  \begin{tabular}{lc}
    \hline
    \textbf{Model} & \textbf{Avg. Recommendation}\\
    \hline
    Llama-3.1-8B-Instruct     & $8.1\pm1.4$       \\
    Llama-3.1-70B-Instruct     &   $6.9\pm2.8$    \\
    Claude-3.5-Sonnet     &      $7.6\pm1.7$    \\
    GPT-4o (2024-11-20)      & $7.7\pm0.8$         \\
    OpenReviewer     & $5.4\pm1.1$        \\
    Human Reviewers & $5.4\pm1.2$ \\
    \hline
  \end{tabular}}
  \caption{Average of recommendations in reviews for 400 test papers, generated with different LLMs and normalized to a scale from 1 (strong reject) to 10 (strong accept).
  The recommendations found in reviews produced by OpenReviewer are much more critical than other LLMs, which tend to give a positive recommendation most of the time.}
  \label{tab:recommendation_avg}
\end{table}

\subsection{Review Arena}
We run an arena-style preference evaluation with an LLM-as-a-judge setup to measure whether OpenReviewer produces better reviews than the other LLMs.
This is similar to MT bench~\citep{mt_bench_llm_judge} and AlpacaEval~\citep{alpaca_eval}, which use an LLM-as-a-judge to evaluate the quality of instruction-tuned language models and chatbots.
While human judgments could be more meaningful, they are expensive to obtain as, in our case, they would require annotators to be trained reviewers and study each paper in great detail.

\begin{figure}[h]
  \includegraphics[width=\columnwidth]{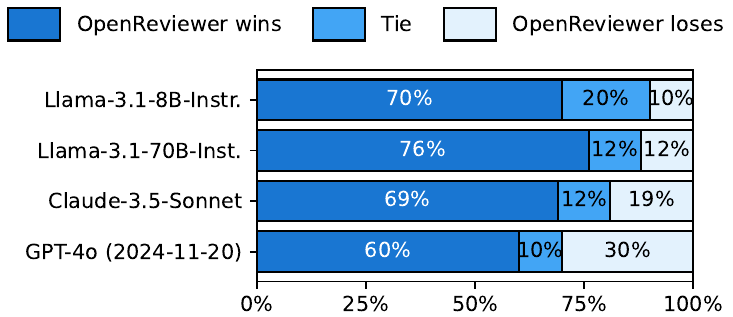}
  \caption{Preference evaluation using GPT-4o as the annotator, judging which generated review aligns better with a set of human-written reviews.}
  \label{fig:winrates}
\end{figure}

Given a set of human ``expert'' reviews and two reviews, A and B, we ask GPT-4o (2024-11-20) to determine whether review A or B aligns better with the given expert reviews.
Specifically, we first ask it to consider how well each section of A and B matches the corresponding section in the expert reviews and then to decide between A, B, or Tie.
Refer to Section~\ref{sec:prompts} for the exact prompts and to Section~\ref{sec:example_outputs} for an example output of the LLM judge.
Figure~\ref{fig:winrates} visualizes the win rates of OpenReviewer against the other LLMs.
According to the LLM judge, OpenReviewer wins against other LLMs for most papers, achieving win rates ranging from $60\%$ against GPT4o to $76\%$ against Llama-3.1-70B-Instruct.

\section{Discussion}
Our results demonstrate that OpenReviewer generates reviews that align considerably better with human expert judgments than general-purpose LLMs. The key findings warrant several important discussions:

\textbf{Review Quality and Criticism Level:} A striking observation is that general-purpose LLMs consistently produce overly positive reviews, with average recommendations between 6.9 and 8.1 on a 10-point scale. In contrast, OpenReviewer's average rating of 5.4 matches the human reviewer distribution exactly. This suggests that specialized training on peer review data helps overcome the tendency of LLMs to be overly favorable - a critical feature for providing constructive feedback that can help improve papers.

\textbf{Use Cases and Limitations:} While OpenReviewer shows promise as a pre-submission feedback tool, it is important to emphasize that it is not intended to replace human peer review. The system can help authors identify potential weaknesses and prepare stronger submissions, but should be used alongside other forms of feedback. Additionally, the model's training data comes primarily from ML/AI venues, potentially limiting its effectiveness for other fields.

\textbf{Ethical Considerations:} The development of automated review systems raises important questions about maintaining review quality and preventing potential misuse. There is a risk that authors might try to ``gam'' such systems or that conferences might be tempted to use them to replace human reviewers. Clear guidelines about appropriate use cases and limitations are essential.

\textbf{Technical Tradeoffs:} Our choice of an 8B parameter model balanced performance with accessibility and computational requirements. While larger models might achieve better results, our evaluation suggests that specialized training on peer review data could be more important than the model scale.
However, we believe scaling up the LLM powering OpenReviewer will further improve the generated reviews.

\section{Related Work}
\subsection{AI-Assisted Peer Review}
Prior work has explored various ways NLP can support the peer review process.
Early approaches focused on reviewer-paper matching using text similarity and topic modeling \citep{toronto_matching,paraphr_sim}.
More recent work has investigated automated analysis of review quality \citep{automate_scientific_reviewing}, detection of biases in peer review \citep{biases_peer_review}, generation of meta-reviews \citep{summarizing_docs_for_metareview}, and manuscript revision~\cite{revise_resubmit}.
ReviewRobot~\citep{review_robot} predicts review scores and generates comments using knowledge graphs.
However, most existing systems target isolated aspects rather than providing comprehensive reviewing support. Our work builds on these efforts while introducing an end-to-end system designed explicitly for generating complete peer reviews.
\citet{what_can_nlp_do_for_peer_review} present an up-to-date overview of opportunities for leveraging natural language processing across the entire peer review process.

\subsection{Review Generation with Large Language Models}
With the emergence of powerful LLMs, several recent studies have explored their potential for automated review generation. 
\citet{reviewer_gpt} conducted experiments using OpenAI's GPT-4 for reviewing papers but found limitations in technical depth and consistency.
\citet{can_llm_provide_useful_feedback} show that while LLMs can provide useful feedback, they tend to focus more on writing and presentation issues rather than technical contributions.
The AI Scientist~\cite{ai_scientist} is a recent framework for fully automated scientific discovery based on GPT4o, including research idea generation, coding, experiment execution, full scientific paper writing, and simulation of the review process.
\citet{lim_gen} investigate approaches to harness open LLMs for producing suggestive limitations of research papers.
\citet{evaluate_novelty} retrieve similar papers to predict the novelty of publications.
Reviewer2~\citep{reviewer2} is a two-stage review generation framework that first produces a set of aspect focused prompts which are then used to generate a review.
SEA~\citep{SEA} is a recent framework for automatic peer reviewing including the standardization, evaluation and analysis of reviews.
Unlike approaches that use general-purpose LLMs, OpenReviewer employs a model specifically fine-tuned on peer review data to better align with reviewing standards and expectations.

\subsection{Datasets and Resources}
Several datasets have been developed to study peer review, including PeerRead \citep{peer_read}, which contains reviews from machine learning conferences, and NLPeer \citep{nlpeer}, which focuses on computational linguistics venues.
MOPRD~\citep{moprd} is a multi-disciplinary dataset for peer review.
Our training data incorporates a larger and more recent collection of human reviews from top ML conferences.

\subsection{Evaluation of Review Quality}
Measuring review quality is an ongoing challenge in the field.
Prior work has proposed various metrics, including the soundness of human reviews \citep{nips_review_process_analysis}, helpfulness ratings \citep{miscalibrations_in_ratings}, and structured quality instruments \citep{review_quality_instrument}.
\citet{llm_reliable_reviewer} evaluate LLM-generated reviews with a question-answering task and find that they are weak in giving critical feedback.
Our evaluation is designed explicitly to assess generated reviews automatically, and we do not use humans-in-a-loop.

\subsection{Ethics and Bias in Peer Review}
Research has highlighted various forms of bias in peer review, including institutional bias \citep{reviewer_bias_single_vs_double}, gender bias \citep{diversity_innovation_paradox}, and strategic manipulation \citep{collusion_rings}.
While automation through NLP offers potential solutions, it also raises new ethical concerns about fairness, transparency, and accountability~\citep{improve_peer_review_nlp,are_we_there_yet,aspr}.

\section{Conclusion}
We presented OpenReviewer, an open-source system for generating high-quality peer reviews of ML/AI papers. Through careful fine-tuning on expert reviews and evaluation against multiple baselines, we demonstrated that OpenReviewer produces more realistic and critical reviews than general-purpose LLMs. Our key contribution is showing that specialized training on peer review data can overcome the tendency of LLMs to generate overly positive assessments.

Looking ahead, several promising directions emerge:

\begin{itemize}
    \item Expanding training data to cover more venues and domains.
    \item Incorporating related literature search and citation graph information to improve the assessment of novelty claims.
    \item Developing better automatic evaluation metrics for review quality.
    \item Creating interfaces for collaborative human-AI reviewing.
\end{itemize}

While OpenReviewer shows promise as a tool for generating preliminary feedback, it should complement rather than replace human peer review. We hope this work spurs further research into AI-assisted academic reviewing while maintaining high standards of scholarly assessment.

\section*{Limitations}
Our study faces several key limitations across data, technical, evaluation, and practical dimensions.
Regarding data, our dataset is restricted to papers from 2022 onwards from only ICLR and NeurIPS conferences within the machine learning and AI domain.
However, given rapid field advances, this temporal bound helps ensure contemporary relevance.
The partial use of non-anonymized papers may also introduce information leakage concerns.
Technical limitations include OpenReviewer's dependence on PDF-to-markdown conversion accuracy and its relatively modest 8B parameter size compared to larger models with potentially better document understanding capabilities.
Our evaluation scope is constrained by a test set of only 400 papers due to commercial LLM usage costs, focuses primarily on automated metrics rather than detailed human analysis, and compares against a limited set of baseline models due to resource constraints.
Practical limitations include performance constraints from running on limited hardware and the risk of giving authors false confidence through automated feedback alone.
Additional challenges involve the bias toward certain conferences and review templates, the limited domain focus on machine learning and AI, the handling of figures and images, and the incorporation of relevant background knowledge from references - areas we continue to work on improving.

\section*{Ethics and Broader Impact Statement}
OpenReviewer raises several important ethical considerations that warrant careful discussion.
While our demo aims to assist authors with pre-submission feedback, it could potentially be misused to automate the peer review process entirely, compromising scientific rigor.
We strongly emphasize that OpenReviewer is designed to complement, not replace, human peer review.

There are also concerns about fairness and bias.
Our training data comes primarily from top ML/AI conferences, which may encode existing biases in the field regarding what constitutes ``good research''.
This could disadvantage work from underrepresented perspectives or methodological approaches.
Additionally, researchers with access to better computational resources might gain an unfair advantage in preparing submissions.

On the positive side, OpenReviewer could democratize access to high-quality feedback, particularly benefiting researchers from institutions without large peer networks or those in resource-constrained environments. Early feedback could help authors improve their work before submission, potentially leading to higher-quality publications and more efficient use of human reviewer time.

To promote responsible use, we:

\begin{itemize}
    \item Explicitly state the system's limitations in the documentation.
    \item Make OpenReviewer and the LLM powering it openly available.
    \item Include clear disclaimers about appropriate use cases.
    \item Encourage further research into bias detection and mitigation.
\end{itemize}
We call on the research community to carefully consider the implications of automated review systems and work together to establish guidelines for their ethical deployment.

\section*{Acknowledgments}
This research was primarily supported by the Leibniz Young Investigator Grant program (project ARENA, LYIG-2023-01) of Leibniz University Hannover, funded by the Ministry of Science and Culture of Lower Saxony (MWK) (grant no. 11-76251-114/2022). Additional computational resources were provided through the AI service center KISSKI (grant no. 01IS22093C), funded by the German Federal Ministry of Education and Research (BMBF).

\bibliography{references}

\appendix

\section{Appendix}
\label{sec:appendix}

\subsection{Prompts}
\label{sec:prompts}
The prompts used by OpenReviewer are shown in Figures 3 and 4.
The prompts used for GPT-4o as a judge are shown in Figures 5 and 6.

\subsection{Example Outputs}
\label{sec:example_outputs}
Figure 7 shows an example output of GPT-4o as a judge, comparing two reviews. 

\subsection{Training Hyperparameters}
Figure 8 shows all training hyperparameters used to train Llama-OpenReviewer-8B with the axolotl library.

\begin{figure*}
\begin{tcolorbox}
\small
You are an expert reviewer for AI conferences. You follow best practices and review papers according to the reviewer guidelines.

Reviewer guidelines:
1. Read the paper: It’s important to carefully read through the entire paper, and to look up any related work and citations that will help you comprehensively evaluate it. Be sure to give yourself sufficient time for this step.
2. While reading, consider the following:
    - Objective of the work: What is the goal of the paper? Is it to better address a known application or problem, draw attention to a new application or problem, or to introduce and/or explain a new theoretical finding? A combination of these? Different objectives will require different considerations as to potential value and impact.
    - Strong points: is the submission clear, technically correct, experimentally rigorous, reproducible, does it present novel findings (e.g. theoretically, algorithmically, etc.)?
    - Weak points: is it weak in any of the aspects listed in b.?
    - Be mindful of potential biases and try to be open-minded about the value and interest a paper can hold for the community, even if it may not be very interesting for you.
3. Answer four key questions for yourself, to make a recommendation to Accept or Reject:
    - What is the specific question and/or problem tackled by the paper?
    - Is the approach well motivated, including being well-placed in the literature?
    - Does the paper support the claims? This includes determining if results, whether theoretical or empirical, are correct and if they are scientifically rigorous.
    - What is the significance of the work? Does it contribute new knowledge and sufficient value to the community? Note, this does not necessarily require state-of-the-art results. Submissions bring value to the community when they convincingly demonstrate new, relevant, impactful knowledge (incl., empirical, theoretical, for practitioners, etc).
4. Write your review including the following information: 
    - Summarize what the paper claims to contribute. Be positive and constructive.
    - List strong and weak points of the paper. Be as comprehensive as possible.
    - Clearly state your initial recommendation (accept or reject) with one or two key reasons for this choice.
    - Provide supporting arguments for your recommendation.
    - Ask questions you would like answered by the authors to help you clarify your understanding of the paper and provide the additional evidence you need to be confident in your assessment.
    - Provide additional feedback with the aim to improve the paper. Make it clear that these points are here to help, and not necessarily part of your decision assessment.

Your write reviews in markdown format. Your reviews contain the following sections:

\verb|#| Review

\verb|{review_fields}|

Your response must only contain the review in markdown format with sections as defined above.
\end{tcolorbox}
\label{fig:openreviewer_system_prompt}
\caption{System prompt used by OpenReviewer. Fields in \{\} are placeholders.}
\end{figure*}

\begin{figure*}
\begin{tcolorbox}
\small
Review the following paper:

\verb|{paper_text}|
\end{tcolorbox}
\label{fig:openreviewer_user_prompt}
\caption{User prompt used by OpenReviewer.  Fields in \{\} are placeholders.}
\end{figure*}

\begin{figure*}
\begin{tcolorbox}
\small
You are an expert meta-reviewer for an AI conference. You will be provided with \verb|{n_expert_reviews}| expert reviews and two additional reviews, review A and review B, all for the same paper. The expert reviews form a groundtruth of reviews. Your task is to determine whether review A or review B aligns better with the given expert reviews.

All reviewers were instructed to write reviews with the following sections:
\verb|{review_fields}|

Think about how well each section of the reviews matches the corresponding section in the expert reviews, except for the summary section. For sections requiring a numerical rating, determine how well the numerical rating matches the numerical ratings of the expert reviews.

All reviews are delimited with XML tags.
Start your response with your thoughts about how well each section of Review A and Review B matches the corresponding section in the expert reviews. Then, provide your decision as either ``Review '', ``Review '', or ``Ti''.
\end{tcolorbox}
\label{fig:llm_judge_system_prompt}
\caption{System prompt for the LLM judge. Fields in \{\} are placeholders.}
\end{figure*}

\begin{figure*}
\begin{tcolorbox}
\small
Expert reviews:
\verb|{expert_reviews}|

Given the expert reviews above, judge which of the following reviews aligns better with the given expert reviews:

\verb|<review_a>|
Review A:
\verb|{review_a}|
\verb|</review_a>|

\verb|<review_b>|
Review B:
\verb|{review_b}|
\verb|</review_b>|
\end{tcolorbox}
\label{fig:llm_judge_user_prompt}
\caption{User prompt for the LLM judge.  Fields in \{\} are placeholders.}
\end{figure*}

\begin{figure*}
\begin{tcolorbox}
\small
\verb|###| Analysis of Review A and Review B in Comparison to Expert Reviews

\verb|####| **Review A**
1. **Soundness**:  
   - Rating: 3  
   - Matches the majority of expert reviews, which rated soundness as 3, except for one expert who rated it as 4. This alignment is reasonable.

2. **Presentation**:  
   - Rating: 3  
   - Matches the expert reviews, which consistently rated presentation as 3.

3. **Contribution**:  
   - Rating: 3  
   - Matches the expert reviews, which consistently rated contribution as 3.

4. **Strengths**:  
   - Review A highlights the clarity of the paper, the theoretical rigor, and the motivation for the problem, which aligns with the expert reviews' emphasis on the novelty of the epigraph form, theoretical guarantees, and empirical results. However, Review A does not mention the gradient conflict resolution or the toy example, which were noted as strengths in the expert reviews.

5. **Weaknesses**:  
   - Review A mentions the focus on tabular MDPs and computational expense, which align with the expert reviews' concerns about computational inefficiency and scalability. However, it does not address the assumptions (e.g., Assumption 2) or the need for more complex real-world applications, which were highlighted in the expert reviews.

6. **Questions**:  
   - The questions in Review A focus on extending the algorithm to continuous spaces, dealing with infinite uncertainty sets, and computational efficiency. These are relevant but do not fully align with the expert reviews, which also asked about the gap between transition kernels, assumptions, and scaling with state space size.

7. **Rating**:  
   - Rating: 6  
   - Matches two of the expert reviews, which rated the paper as 6. The other two expert reviews rated it as 8 and 3, so this rating is within the range of expert opinions.

\verb|####| **Review B**
1. **Soundness**:  
   - Rating: 4  
   - Matches one expert review that rated soundness as 4, but is higher than the other three expert reviews, which rated it as 3. This rating is slightly optimistic compared to the majority of expert reviews.

2. **Presentation**:  
   - Rating: 3  
   - Matches the expert reviews, which consistently rated presentation as 3.

3. **Contribution**:  
   - Rating: 4  
   - Higher than the expert reviews, which consistently rated contribution as 3. This rating is more optimistic than the expert consensus.

4. **Strengths**:  
   - Review B emphasizes the novelty of the epigraph form, theoretical rigor, and empirical validation, which align well with the expert reviews. It also mentions the clarity in problem formulation and the generality of the approach, which are consistent with the expert reviews. However, it goes further by claiming the contribution is "substantial" and "impactful," which is more enthusiastic than the expert reviews.

5. **Weaknesses**:  
   - Review B mentions the complexity of presentation, computational efficiency, limited experimental scope, and lack of comparison to alternative methods. These align partially with the expert reviews, which also noted computational inefficiency and scalability issues. However, Review B does not address the assumptions or the need for more complex real-world applications, which were key weaknesses in the expert reviews.

6. **Questions**:  
   - The questions in Review B focus on scalability, intuition for the epigraph form, scenarios where the Lagrangian approach might be preferable, and sensitivity to hyperparameters. These are relevant but do not fully align with the expert reviews, which also asked about assumptions, scaling with state space size, and the gap between transition kernels.

7. **Rating**:  
   - Rating: 8  
   - Matches one expert review that rated the paper as 8, but is higher than the other three expert reviews, which rated it as 6, 6, and 3. This rating is more optimistic than the majority of expert opinions.

---

\verb|###| Decision
Both reviews align with the expert reviews to some extent, but Review A aligns more closely overall:
- Review A's numerical ratings for soundness, presentation, and contribution match the majority of expert reviews, while Review B's ratings for soundness and contribution are more optimistic.
- Review A's strengths and weaknesses align more closely with the expert reviews, though it misses some details (e.g., assumptions). Review B is more enthusiastic and introduces additional points (e.g., generality, comparison to alternatives) that are not emphasized in the expert reviews.
- Both reviews' questions are relevant but do not fully align with the expert reviews.

**Final Decision**: **Review A** aligns better with the expert reviews.

\end{tcolorbox}
\label{fig:llm_judge_output}
\caption{Example output from the LLM judge. Review A is generated by OpenReviewer, and Review B is from GPT-4o.}
\end{figure*}

\label{sec:hyperparams}
\begin{figure*}
\small
\begin{verbatim}
base_model: meta-llama/Llama-3.1-8B-Instruct

plugins:
  - axolotl.integrations.liger.LigerPlugin
liger_rope: true
liger_rms_norm: true
liger_glu_activation: true
liger_fused_linear_cross_entropy: true
strict: false
chat_template: llama3
datasets:
  - path: ""
    type:
    ds_type: parquet
sequence_len: 131072 
sample_packing: false
pad_to_sequence_len: true
gradient_accumulation_steps: 1
micro_batch_size: 1
num_epochs: 3
optimizer: adamw_torch
lr_scheduler: cosine
learning_rate: 2e-5
train_on_inputs: false
group_by_length: false
bf16: auto
tf32: true
gradient_checkpointing: true
logging_steps: 1
flash_attention: true
warmup_steps: 50
deepspeed: deepspeed_configs/zero3_bf16.json
weight_decay: 0.0
fsdp:
fsdp_config:
special_tokens:
  pad_token: <|finetune_right_pad_id|>
  eos_token: <|eot_id|>
\end{verbatim}
\label{fig:hyperparams}
\caption{The axolotl configuration used to train Llama-OpenReviewer-8B, including all hyperparameters.}
\end{figure*}

\end{document}